\documentclass{ieeeaccess}
\usepackage[pdftex]{graphicx}
\usepackage{url}
\usepackage{multirow}
\usepackage{times}
\usepackage[labelformat=simple]{subcaption}

\usepackage{cite}
\usepackage{amsmath,amssymb,amsfonts}
\usepackage{algorithmic}
\usepackage{textcomp}
\def\BibTeX{{\rm B\kern-.05em{\sc i\kern-.025em b}\kern-.08em
    T\kern-.1667em\lower.7ex\hbox{E}\kern-.125emX}}

\usepackage{color}
\definecolor{orange}{rgb}{1,.5,0}
\definecolor{purple}{rgb}{1,0,1}
\definecolor{gray}{rgb}{.9,.9,.9}
\definecolor{brown}{rgb}{0.59, 0.29, 0.0}

\newcommand{\lw}[1]{\smash{\lower2.0ex\hbox{#1}}}
\newcommand{\lwm}[1]{\smash{\lower1.5ex\hbox{#1}}}
\newcommand{\lwh}[1]{\smash{\lower1.0ex\hbox{#1}}}

\newcommand{\xmark}{$\times$}

\newcounter{mycounterA} % カウンタの宣言
\renewcommand\themycounterA{\Alph{mycounterA}}
\newcommand{\useMycounterA}[1][]{\refstepcounter{mycounterA}{#1}{\themycounterA}}

\newcounter{mycountera} % カウンタの宣言
\renewcommand\themycountera{\alph{mycountera}}
\newcommand{\useMycountera}[1][]{\refstepcounter{mycountera}{#1}{\themycountera}}

\graphicspath{{fig/}}

\begin{document}

\history{Date of publication xxxx 00, 0000, date of current version xxxx 00, 0000.}
\doi{10.1109/ACCESS.2019.2960566 }

\title{ShakeDrop Regularization for Deep Residual Learning}

\author{\uppercase  {Yoshihiro Yamada}\authorrefmark{1}, \uppercase{Masakazu Iwamura}\authorrefmark{1},
\IEEEmembership  {Member, IEEE}, \uppercase{Takuya Akiba}\authorrefmark{2}, \uppercase{ and Koichi Kise}\authorrefmark{1},
\IEEEmembership{Member, IEEE}}
\address[1]{Graduate School of Engineering, Osaka Prefecture University, Sakai, Osaka 599-8531, Japan}
\address[2]{Preferred Networks, Inc., Chiyoda, Tokyo
100-0004, Japan}

\tfootnote{This work was supported by JST CREST JPMJCR16E1, JSPS KAKENHI JP25240028, JP17H01803, JP18J15255 and JP18K19785, JST AIP PRISM J18ZZ00418, the Artificial Intelligence Research Promotion Foundation, and the AWS Cloud Credits for Research Program. © 2019 IEEE.  Personal use of this material is permitted.  Permission from IEEE must be obtained for all other uses, in any current or future media, including reprinting/republishing this material for advertising or promotional purposes, creating new collective works, for resale or redistribution to servers or lists, or reuse of any copyrighted component of this work in other works.}
\markboth
{Author \headeretal: Preparation of Papers for IEEE TRANSACTIONS and JOURNALS}
{Author \headeretal: Preparation of Papers for IEEE TRANSACTIONS and JOURNALS}
\corresp{Corresponding author: Yoshihiro Yamada (e-mail: yamada@m.cs.osakafu-u.ac.jp).}

\begin{abstract}

Overfitting is a crucial problem in deep neural networks, even in the latest network architectures.
In this paper, to relieve the overfitting effect of ResNet and its improvements (i.e., Wide ResNet, PyramidNet, and ResNeXt), we propose a new regularization method called ShakeDrop regularization.
ShakeDrop is inspired by Shake-Shake, which is an effective regularization method, but can be applied to ResNeXt only.
ShakeDrop is more effective than Shake-Shake and can be applied not only to ResNeXt but also ResNet, Wide ResNet, and PyramidNet.
An important key is to achieve stability of training.
Because effective regularization often causes unstable training, we introduce a training stabilizer, which is an unusual use of an existing regularizer.
Through experiments under various conditions, we demonstrate the conditions under which ShakeDrop works well. 

\end{abstract}


\titlepgskip=-15pt

\maketitle

\section{Introduction}
\label{sec:introduction01}

Recent advances in generic object recognition have been achieved using  deep neural networks.
Since ResNet~\cite{He_CVPR2016} created the opportunity to use very deep convolutional neural networks (CNNs) of over a hundred layers by introducing the building block, its improvements, such as Wide ResNet~\cite{Zagoruyko2016WRN}, PyramidNet~\cite{Han_CVPR2017,arXiv:1610.02915}, and ResNeXt~\cite{Xie_CVPR2017} have broken records for the lowest error rates.

The development of such base network architectures, however, is not sufficient to reduce the generalization error (i.e., difference between the training and test errors)
due to over-fitting. In order to improve test errors, regularization methods which are  processes to introduce additional information to CNNs have been proposed~\cite{Miyato_TPAMI2018}.
%It has been reported that some regularization methods reduce the generalization error.
Widely used regularization methods include data augmentation~\cite{NIPS2012_4824}, stochastic gradient descent (SGD)~\cite{Zhang_ICLR2017}, weight decay~\cite{NIPS1991_563}, batch normalization (BN)~\cite{Ioffe_ICML2015}, label smoothing~\cite{DBLP:conf/cvpr/SzegedyVISW16}, adversarial training~\cite{43405}, mixup~\cite{zhang2018mixup, Tokozume_2018_CVPR, Verma2018_arXiv, Takahashi_ACML2018}, and dropout~\cite{Srivastava:2014:DSW:2627435.2670313, pmlr-v28-wan13}.
Because the generalization errors when regularization methods are used are still large, effective regularization methods have been studied.

%\red{Because of historical background and implementation differences, various regularization methods have been proposed~\cite{Zhang_ICLR2017}, and the effectiveness has been evaluated by the generalization errors quantitatively.}
Recently, an effective regularization method which achieved the lowest test error called Shake-Shake regularization~\cite{Gastaldi_ICLRW2017,arXiv:1705.07485} was proposed.
It is an interesting method, which, in training, disturbs the calculation of the forward pass using a random variable, and also that of the backward pass using a different random variable.
Its effectiveness was proven by an experiment on ResNeXt, to which Shake-Shake was applied (hereafter, this type of combination is denoted by ``ResNeXt + Shake-Shake''), which achieved the lowest error rate on CIFAR-10/100 datasets~\cite{CIFAR-10_100}.
Shake-Shake, however, has the following two drawbacks:
(i)~it can be applied to ResNeXt only, and
(ii)~the reason it is effective has not yet been identified.

The current paper addresses these problems.
For problem (i),  we propose a novel powerful regularization method called \textit{ShakeDrop regularization}, which is more effective than Shake-Shake.
Its main  advantage is that it has the potential to be applied not only to ResNeXt (hereafter, three-branch architectures) but also ResNet, Wide ResNet, and PyramidNet (hereafter, two-branch architectures).
The main difficulty to overcome is unstable training .
We solve this problem by proposing a new stabilizing mechanism for \textit{difficult-to-train} networks.
For problem (ii), in the process of deriving ShakeDrop, we provide an intuitive interpretation of Shake-Shake.
Additionally, we present the mechanism in which ShakeDrop works.
Through experiments using various base network architectures and parameters, we demonstrate the conditions under which ShakeDrop successfully works. 

This paper is an extended version of ICLR workshop paper~\cite{Yamada_ICLRW2018}.

\begin{figure*}[tb]
%\centering
  %\begin{tabular}{c}
  \begin{minipage}{0.46\hsize}
    \centering
    %\fbox{ 
    \includegraphics[width=\hsize]{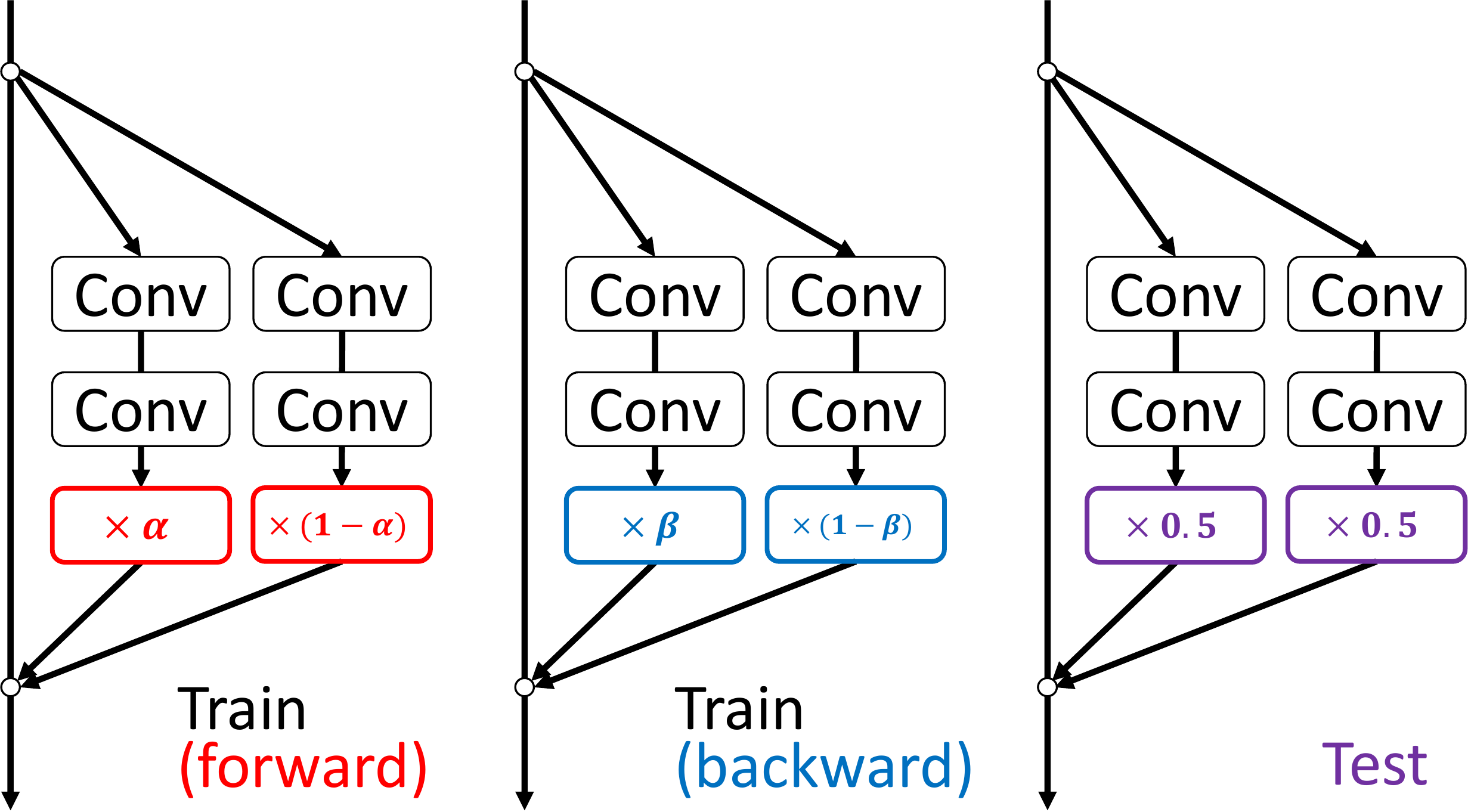}
    %}
    \subcaption{Shake-Shake only for ResNeXt~\cite{arXiv:1705.07485}}
    \label{fig:shakeshake}
  \end{minipage}
  \hspace{0.04\hsize}
  \begin{minipage}{0.46\hsize}
    \centering
    %\fbox{ 
    \includegraphics[width=\hsize]{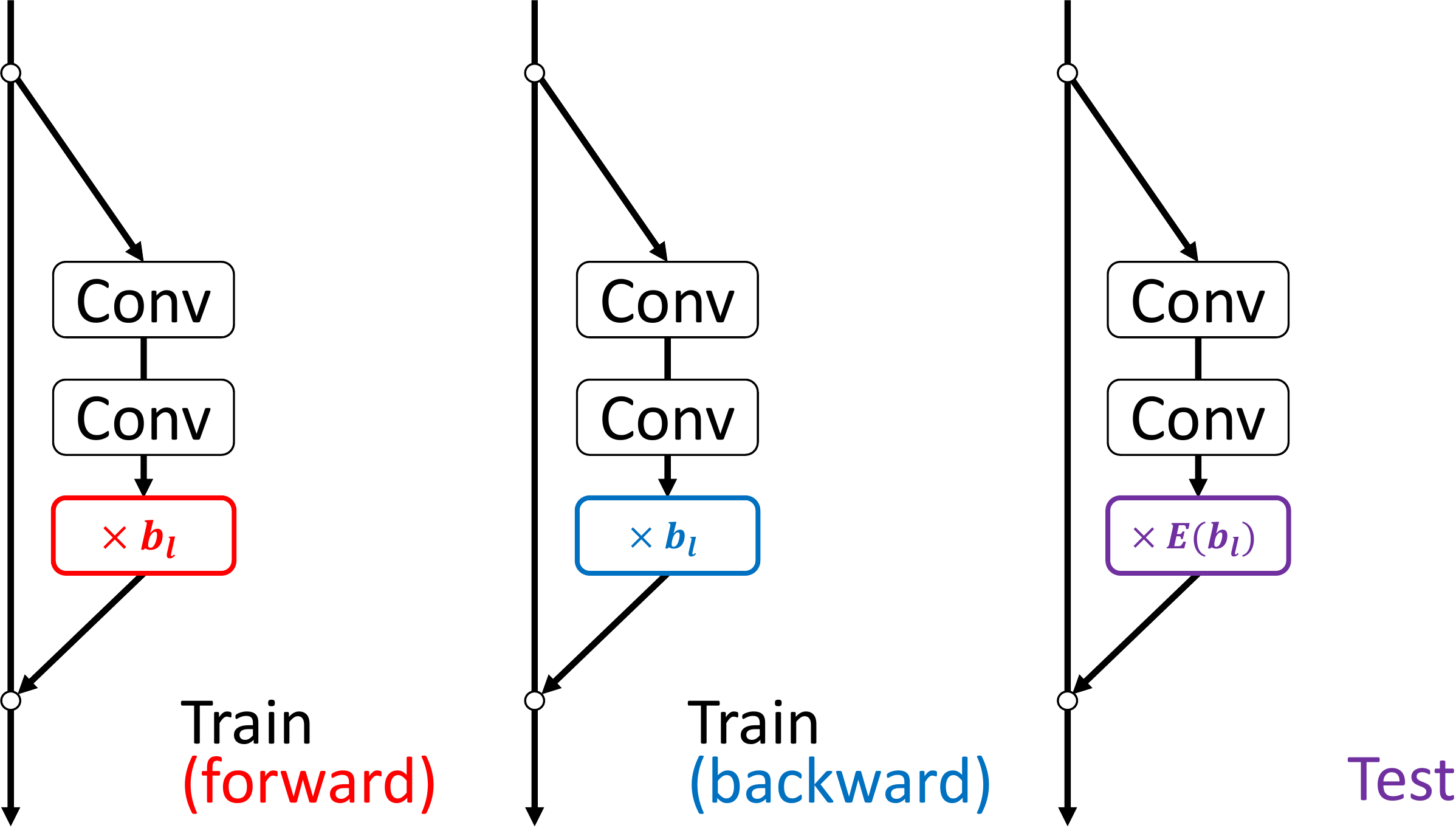}
    %}
    \subcaption{RandomDrop for the two-branch ResNet family~\cite{arXiv:1603.09382}}
    \label{fig:randomdrop}
  \end{minipage}
  \hspace{0.04\hsize}
  \begin{minipage}{0.46\hsize}
    \centering
    %\fbox{ 
    \includegraphics[width=\hsize]{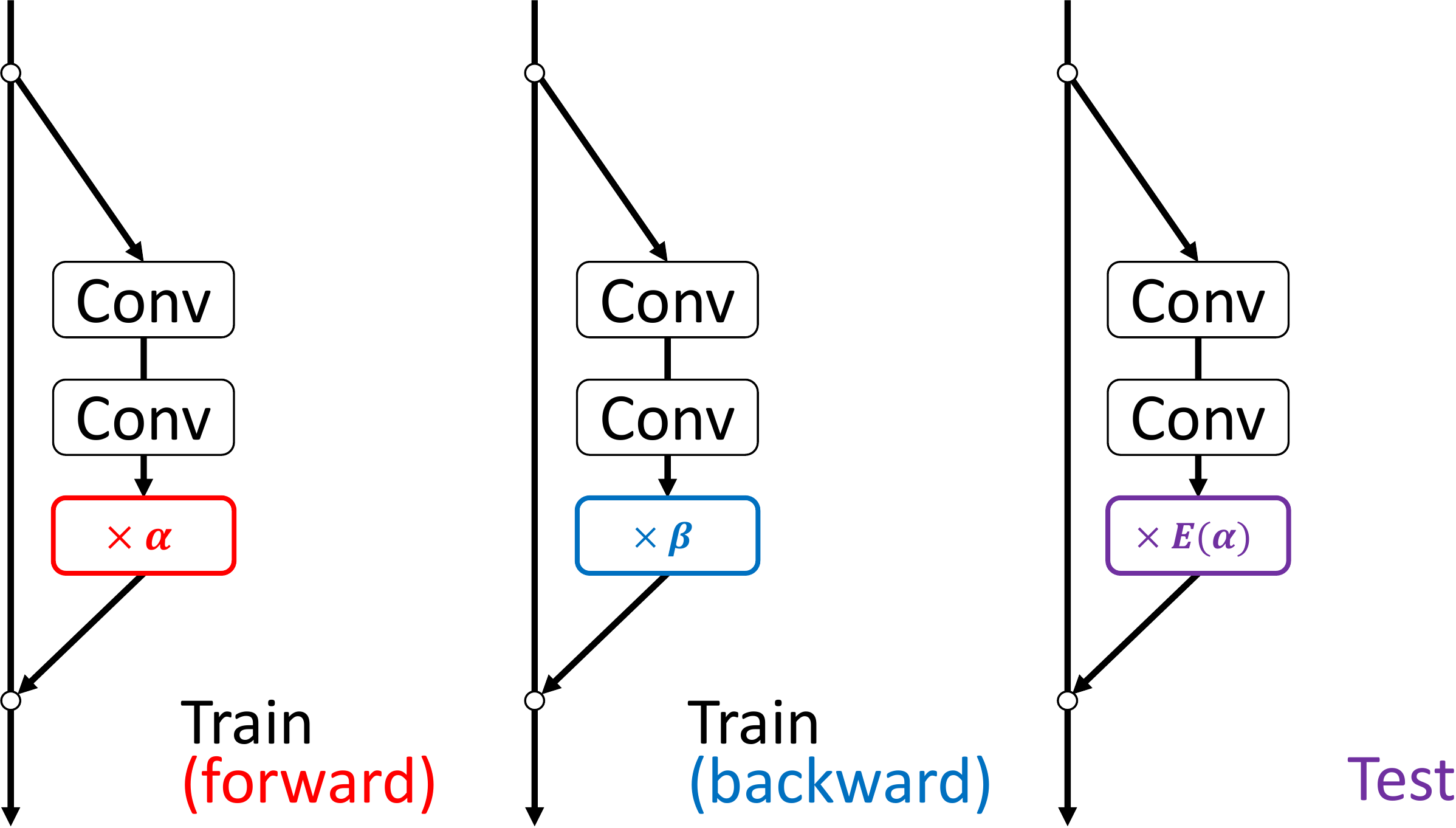}
    %}
    \subcaption{[Intermediate method] \textit{Single-branch Shake}}
    \label{fig:pyramidshake}
  \end{minipage}
  \hspace{0.04\hsize}
  \begin{minipage}{0.46\hsize}
    \centering
    %\fbox{ 
    \includegraphics[width=\hsize]{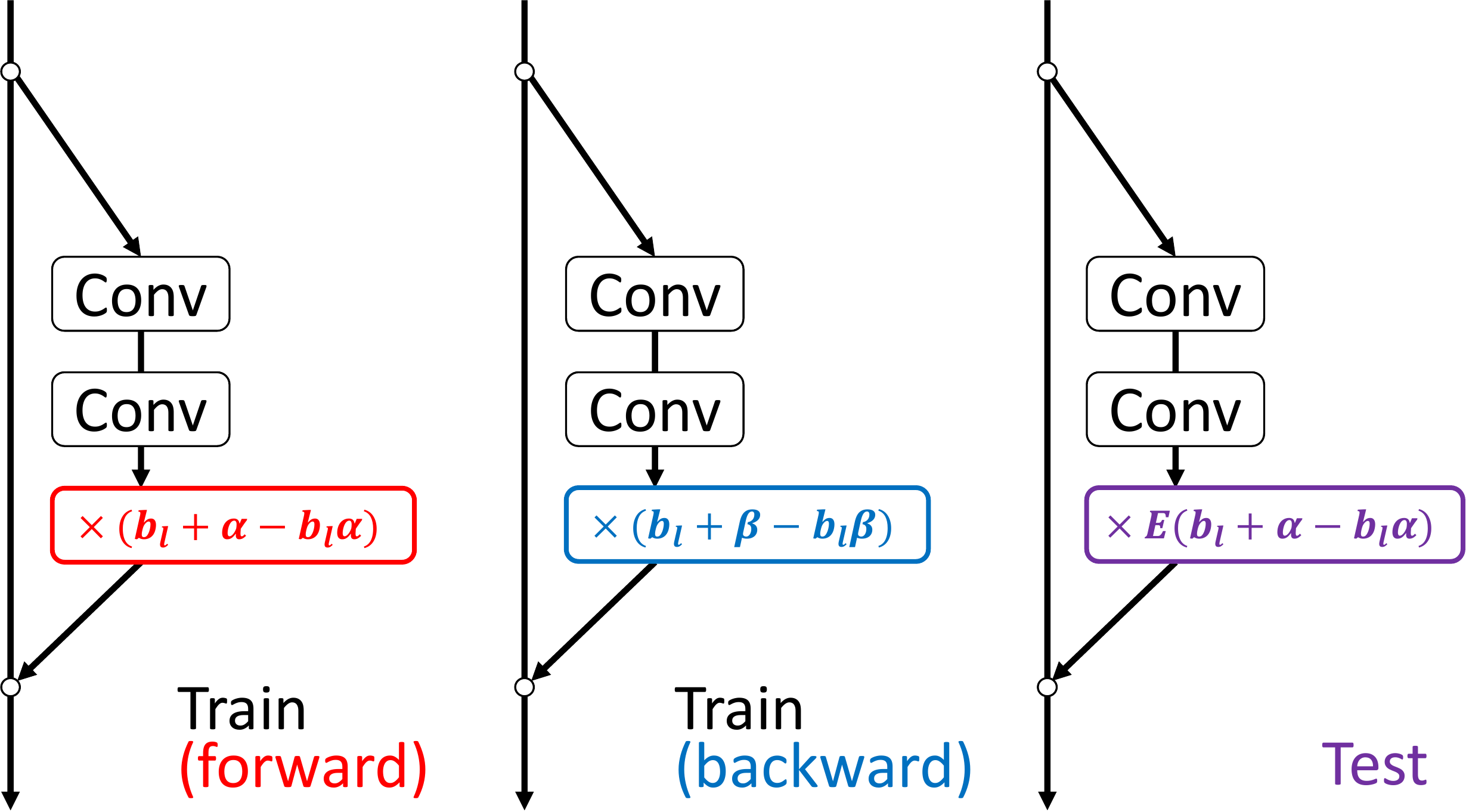}
    %}
    \subcaption{ShakeDrop for the two- and  three-branch ResNet family}
    \label{fig:shakedrop}
  \end{minipage}
  \hspace{0.05\hsize}
  \caption{Regularization methods for the ResNet family.
  (a) and (b) are existing methods.
  (c) is an intermediate regularization method used to derive  the proposed method.
  (d) is the proposed method.
  ``Conv'' denotes a convolution layer; $E[x]$ denotes the expected value of $x$; and $\alpha$, $\beta$, and $b_l$ denote random coefficients.}
  \label{fig:architecture}
  %\end{tabular}
\end{figure*}

\section{Regularization methods for the ResNet family}
\label{subsec:regularization}

In this section, we present two regularization methods for the ResNet family,
both of which are used to derive the proposed method.\\

\noindent
\textbf{Shake-Shake regularization}~\cite{Gastaldi_ICLRW2017,arXiv:1705.07485} is an effective regularization method for ResNeXt.
It is illustrated in Fig.~\ref{fig:shakeshake}.
The basic ResNeXt building block, which has a three-branch architecture, is given as
\begin{equation}
\label{eqn:resnext}
	G(x) = x + F_1(x) + F_2(x),
\end{equation}
where $x$ and $G(x)$ are the input and output of the building block, respectively, and $F_1(x)$ and $F_2(x)$ are the outputs of two residual branches.

Let $\alpha$ and $\beta$ be independent random coefficients uniformly drawn  from the uniform distribution on the interval $[0,1]$. Then Shake-Shake is given as
\begin{align}
\label{eqn:shakeshake}
  G(x) =
  \begin{cases}
    x + \alpha F_1(x) + (1-\alpha) F_2(x), & \textrm{in~train-fwd} \\
    x + \beta F_1(x) + (1-\beta) F_2(x), & \textrm{in~train-bwd} \\
    x + E[\alpha] F_1(x) + E[1-\alpha] F_2(x), & \textrm{in~test,}
  \end{cases}
\end{align}
where \textit{train-fwd} and \textit{train-bwd} denote the forward and backward passes of training, respectively.
Expected values
$E[\alpha]=E[1-\alpha]=0.5$.
Equation ~\eqref{eqn:shakeshake} means that the calculation of the forward pass is multiplied by random coefficient $\alpha$ and that of the backward pass by another random coefficient $\beta$.
The values of $\alpha$ and $\beta$ are drawn for  each image or batch.
In this paper, we suggest training for longer than usual (more precisely, six times as long as usual).

In the training of neural networks, if the output of a residual branch is multiplied by coefficient $\alpha$ in the forward pass, then it is natural to multiply the gradient by the same coefficient (i.e., $\alpha$) in the backward pass.
Hence, compared with the standard approach, Shake-Shake makes the gradient $\beta/\alpha$ times as large as the correctly calculated gradient on one branch and $(1-\beta)/(1-\alpha)$ times on the other branch.
It seems that the disturbance prevents the network parameters from being captured in local minima.
However, the reason why such a disturbance is effective has not been sufficiently identified.
\\[.1em]

\noindent
\textbf{RandomDrop regularization (a.k.a., Stochastic Depth and ResDrop)}~\cite{DBLP:conf/eccv/HuangSLSW16} is a regularization method originally proposed for ResNet, and also applied to PyramidNet~\cite{arXiv:1612.01230}.
It is illustrated in Fig.~\ref{fig:randomdrop}.
The basic ResNet building block, which has a two-branch architecture, is given as 
\begin{equation}
\label{eqn:resnet_building_block}
	G(x) = x + F(x),
\end{equation}
where $F(x)$ is the output of the residual branch.
RandomDrop makes the network appear to be shallow in learning by dropping some stochastically selected building blocks. 
The $l^\textrm{th}$ building block from the input layer is given as
\begin{equation}
\label{eqn:resdrop}
	G(x) =
  \begin{cases}
    x + b_l F(x), & \textrm{in~train-fwd} \\
    x + b_l F(x), & \textrm{in~train-bwd} \\
    x + E[b_l] F(x), & \textrm{in~test,}
  \end{cases}
\end{equation}
where $b_l \in \{0,1\}$ is a Bernoulli random variable with the probability $P(b_l=1)=E[b_l]=p_l$.
In this paper, we recommend the \textit{linear decay rule} to determine $p_l$, which is given as
\begin{equation}
\label{eqn:linear_decay_rule}
	p_l = 1- \frac{l}{L}(1-p_L),
\end{equation}
where $L$ is the total number of   building blocks and $p_L$ is the initial parameter.
We suggest using $p_L = 0.5$.

RandomDrop can be regarded as a simplified version of dropout~\cite{Srivastava:2014:DSW:2627435.2670313}.
The main difference is that RandomDrop drops layers, whereas dropout drops elements. 

\section{Proposed Method}

\subsection{ShakeDrop Regularization}
\label{subsec:ShakeDrop}

The proposed \textit{ShakeDrop}, illustrated in Fig.~\ref{fig:shakedrop}, is given as
\begin{equation}
\label{eqn:ShakeDrop}
	G(x) =
  \begin{cases}
    x + (b_l+\alpha-b_l \alpha) F(x), & \textrm{in~train-fwd} \\
    x + (b_l+\beta-b_l \beta) F(x), & \textrm{in~train-bwd} \\
    x + E[b_l+\alpha-b_l \alpha] F(x), & \textrm{in~test,}
  \end{cases}
\end{equation}
where $b_l$ is a Bernoulli random variable with probability $P(b_l=1)=E[b_l]=p_l$ given by the linear decay rule \eqref{eqn:linear_decay_rule} in each layer, and $\alpha$ and $\beta$ are independent uniform random variables in each element.
The most effective ranges of $\alpha$ and $\beta$ were experimentally found to be different from those of Shake-Shake, and  are $\alpha = 0$, $\beta \in [0,1]$ and $\alpha \in [-1, 1]$, $\beta \in [0,1]$.
Further details of the parameters are presented in Sections~\ref{sec:preliminary_experiments} and \ref{sec:experiments}.

In the training phase, $b_l$ controls the behavior of ShakeDrop.
If $b_l=1$, then \eqref{eqn:ShakeDrop} is deformed as
\begin{equation}
\label{eqn:ShakeDrop_bl=1}
  G(x) = 
  \begin{cases}
    x + F(x), &   \textrm{in~train-fwd} \\
    x + F(x), &   \textrm{in~train-bwd};\\
  \end{cases}
\end{equation}
that is, ShakeDrop is equivalent to the original network (e.g., ResNet).
If $b_l=0$, then \eqref{eqn:ShakeDrop} is deformed as
\begin{equation}
\label{eqn:ShakeDrop_bl=0}
  G(x) = 
  \begin{cases}
    x + \alpha F(x), &  \textrm{in~train-fwd} \\
    x + \beta F(x), &  \textrm{in~train-bwd}; \\
  \end{cases}
\end{equation}
that is, the calculation of $F(x)$ is perturbed by $\alpha$ and $\beta$.

\subsection{Derivation of ShakeDrop} 
\label{subsec:derivation}

\subsubsection{Interpretation of Shake-Shake regularization}
\label{subsubsec:interpretation_shake-shake}

We provide an intuitive interpretation of Shake-Shake;
to the best of our knowledge, it has not been provided yet.
As shown in \eqref{eqn:shakeshake}
(and in Fig.~\ref{fig:shakeshake}),
in the forward pass, Shake-Shake interpolates the outputs of two residual branches (i.e., $F_1(x)$ and $F_2(x)$) with random weight $\alpha$.
DeVries and Taylor~\cite{DeVries_ICLRW2017} demonstrated that the interpolation of two data in the feature space can synthesize reasonable augmented data; hence
the interpolation in the forward pass of Shake-Shake can be interpreted as synthesizing reasonable augmented data. 
The use of random weight $\alpha$ enables us to generate many different augmented data.
By contrast , in the backward pass, a different random weight $\beta$ is used to disturb the updating parameters, which is expected to help to prevent parameters from being caught in local minima by enhancing the effect of SGD~\cite{Kleinberg_ICML2018}.

\subsubsection{Single-branch Shake Regularization}
\label{subsubsec:1branchshake}

The regularization mechanism of Shake-Shake relies on two or more residual branches; hence, it can only be applied to three-branch network architectures (i.e., ResNeXt).
To achieve a similar regularization to Shake-Shake on two-branch architectures (i.e., ResNet, Wide ResNet, and PyramidNet), we need a different mechanism from interpolation in the forward pass that can synthesize augmented data in the feature space.
In fact, DeVries and Taylor~\cite{DeVries_ICLRW2017} demonstrated not only interpolation but also noise addition in the feature space, which generates reasonable augmented data.
Hence, following Shake-Shake, we apply random perturbation to the output of a residual branch (i.e., $F(x)$ of \eqref{eqn:resnet_building_block});
that is, it is given as
\begin{equation}
\label{eqn:PS}
	G(x) =
  \begin{cases}
    x + \alpha F(x), & \textrm{in~train-fwd} \\
    x + \beta F(x), & \textrm{in~train-bwd} \\
    x + E[\alpha] F(x), & \textrm{in~test.}
  \end{cases}
\end{equation}
We call this regularization method \textit{Single-branch Shake}.
It is illustrated in Fig.~\ref{fig:pyramidshake}.
\textit{Single-branch Shake} is expected to be as effective as Shake-Shake. 
However, it does not work well in practice.
For example, in our preliminary experiments, we applied it to 110-layer PyramidNet with $\alpha \in [0,1]$ and $\beta \in [0,1]$ following Shake-Shake. However, the result on the CIFAR-100 dataset was significantly  bad (i.e., an error rate of 77.99\%).

\begin{table*}[tbp]
  \centering
  \caption{Regularization methods that generate new data.
  ``Sample-wise generation'' means that data is generated using a single sample.
  }
  \label{tab:data_augmentation_summary}
    \begin{tabular}{c||c|c|c|c} \hline
    \lwm{Regularization method} & \multicolumn{3}{c|}{Data augmentation} & Sample-wise \\ \cline{2-4}
      & In (input) data space & In feature space & In label space & generation \\ \hline \hline
      Data augmentation~\cite{NIPS2012_4824} & \checkmark & & & \checkmark \\ \hline
      Adversarial training~\cite{43405} & \checkmark & & & \checkmark \\ \hline
      Label smoothing~\cite{DBLP:conf/cvpr/SzegedyVISW16} & & &\checkmark & \checkmark \\ \hline
      Mixup~\cite{zhang2018mixup, Tokozume_2018_CVPR, Takahashi_ACML2018} & \checkmark & & \checkmark & \\ \hline
      Manifold mixup~\cite{Verma2018_arXiv} & \checkmark & \checkmark & \checkmark & \\ \hline \hline
      Shake-Shake~\cite{Gastaldi_ICLRW2017,arXiv:1705.07485} & & \checkmark & & \checkmark \\ \hline
      ShakeDrop  & & \checkmark & & \checkmark \\ \hline
    \end{tabular}
\end{table*}

\subsubsection{Stabilization of training}
\label{subsubsec:stabilization}

In this section, we consider what caused the failure of \textit{Single-branch Shake}. 
A natural guess is that Shake-Shake has a stabilizing mechanism that \textit{Single-branch Shake} does not have.
The mechanism is ``two residual branches.''
We present an argument to verify whether this is the case. 
As presented in Section~\ref{subsec:regularization}, in training, Shake-Shake makes the gradients of two branches $\beta/\alpha$ times and $(1-\beta)/(1-\alpha)$ times as large as the correctly calculated gradients.
Thus, when $\alpha$ is close to zero or one, it cannot converge (ruin) training because it could make a gradient prohibitively large\footnote{This idea is supported by an experiment that limited the ranges of $\alpha$ and $\beta$ in Shake-Shake~\cite{arXiv:1705.07485}.
When $\alpha$ and $\beta$ were kept close (more precisely, on the number line, $\alpha$ and $\beta$ were on the same side of 0.5, such as $\alpha=0.1$ and $\beta=0.2$), Shake-Shake achieved relatively high accuracy.
However, when $\alpha$ and $\beta$ were kept far apart ($\alpha$ and $\beta$ were on the opposite sides of 0.5, such as $\alpha=0.1$ and $\beta=0.7$), the accuracy was relatively low. 
This indicates that when $\beta/\alpha$ or $(1-\beta)/(1-\alpha)$ were large, training could become less stable.}.
However, two residual branches of Shake-Shake work as a fail-safe system;
that is, even if the coefficient on one branch is large, the other is kept small.
Hence, training on at least one branch is not ruined.
\textit{Single-branch Shake}, however, does not have such a fail-safe system.

From the discussion above, the failure of \textit{Single-branch Shake} was caused by the perturbation being too strong  and the lack of a stabilizing mechanism.
Because weakening the perturbation would just weaken the effect of regularization, we need a method  to stabilize unstable learning under strong perturbation.

\begin{figure}[tb]
    \includegraphics[width=\hsize]{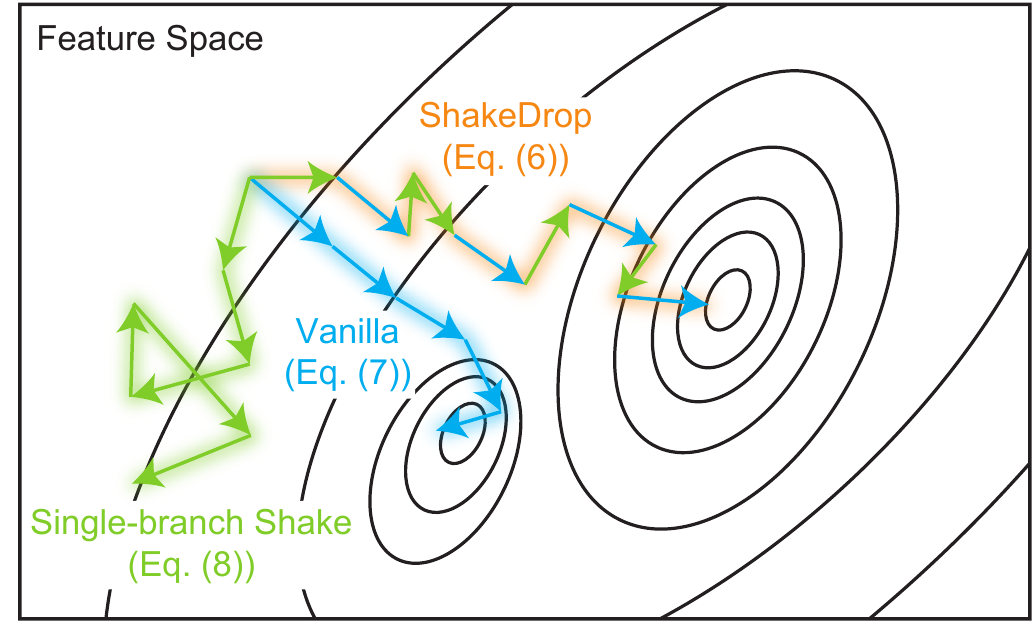}
    \caption{Conceptual sketch of converging trajectories. The original networks called Vanilla \eqref{eqn:ShakeDrop_bl=1} can converge but become trapped in local minima. Single-branch Shake \eqref{eqn:ShakeDrop_bl=0}, which updates the parameters with the strong perturbation, does not become trapped in local minima but cannot converge. Using the probabilistic switches of \eqref{eqn:ShakeDrop_bl=1} and \eqref{eqn:ShakeDrop_bl=0}, ShakeDrop is expected to not become in local minima and to converge to a better minimum.}
    \label{fig:convergence}
    \vspace*{-3mm}
\end{figure}

We propose using the mechanism of RandomDrop to solve the issue.
RandomDrop is designed to make a network apparently shallow to avoid the problems of vanishing gradients, diminishing feature reuse, and a long training time.
In our scenario, the original use of RandomDrop does not have a positive effect because a shallower version of a strongly perturbed network (e.g., a shallow version of ``PyramidNet + \textit{Single-branch Shake}'') would also suffer from strong perturbation.
Thus, we use the mechanism of RandomDrop as a probabilistic switch for  the following two network architectures:
\begin{enumerate}
    \vspace{-0.2\baselineskip}
    \setlength{\itemsep}{1pt}

\item the original network (e.g., PyramidNet), which corresponds to \eqref{eqn:ShakeDrop_bl=1}, and

\item a network that suffers from strong perturbation (e.g., ``PyramidNet + \textit{Single-branch Shake}''), which corresponds to \eqref{eqn:ShakeDrop_bl=0}.
    \vspace{-0.2\baselineskip}
\end{enumerate}
By mixing them up, as shown in Fig.~\ref{fig:convergence}, it is expected that
(i)~when the original network is selected, learning is correctly promoted, and
(ii)~when the network with strong perturbation is selected, learning is disturbed.

To achieve good performance, the two networks should be well balanced, which is controlled by parameter $p_L$.
We discuss  this issue in Section~\ref{sec:preliminary_experiments}.

\subsection{Relationship with existing regularization methods}
\label{subsec:relationship_existing_methods}

In this section, we discuss the relationship between ShakeDrop and existing regularization methods.
Among them, SGD and weight decay are commonly used techniques in the training of deep neural networks.
Although they were not designed for regularization, researchers have indicated that  they have generalization effects~\cite{Zhang_ICLR2017,NIPS1991_563}.
BN~\cite{Ioffe_ICML2015} is a strong regularization technique that has been widely used in recent network architectures.
ShakeDrop is appended to these regularization methods.

ShakeDrop differs from RandomDrop~\cite{DBLP:conf/eccv/HuangSLSW16} and dropout~\cite{Srivastava:2014:DSW:2627435.2670313,pmlr-v28-wan13} in the following two ways:
they do not explicitly generate new data and
they do not update network parameters based on noisy gradients. ShakeDrop coincides with RandomDrop when $\alpha=\beta=0$ instead of the recommended parameters.

Some methods regularize by generating new data.
They are summarized in Table~\ref{tab:data_augmentation_summary}.
Data augmentation~\cite{NIPS2012_4824} and adversarial training~\cite{43405} synthesize data in the (input) data space.
They differ in how they generate data.
The former uses manually designed means, such as random crop and horizontal flip, whereas the latter automatically generates data that should be used for training to improve generalization performance.
Label smoothing~\cite{DBLP:conf/cvpr/SzegedyVISW16} generates (or changes) labels for existing data.
The methods mentioned above generate new data using a single sample.
By contrast, some methods require multiple samples to generate new data.
Mixup~\cite{zhang2018mixup},  BC learning~\cite{Tokozume_2018_CVPR}, and RICAP~\cite{Takahashi_ACML2018} generate new data and their corresponding class labels by interpolating two or more data.
Although they generate new data in the data space, manifold mixup~\cite{Verma2018_arXiv} also does it in the feature space.
Compared with ShakeDrop, which generates data in the feature space using a single sample, none of these regularization methods are in the same category, except for Shake-Shake.

Note that the selection of regularization methods is not always exclusive.
We have successfully used ShakeDrop combined with mixup (see Section~\ref{sec:mixup}).
Although regularization methods in the same category may not be used together (e.g., ``mixup and BC learning'' and ``ShakeDrop and Shake-Shake''), those of different categories may be used together.
Thus, developing the best method in a category is meaningful.

\begin{table*}[tbp]
  \centering
  \caption{\textbf{[Ranges of $\alpha$ and $\beta$]} Average top-1 errors (\%) of ``ResNet + ShakeDrop,'' ``ResNet (EraseReLU version) + ShakeDrop,'' and ``PyramidNet + ShakeDrop'' of four runs at the final (300th) epoch on the CIFAR-100 dataset using  the batch-level update rule.
  ``\xmark'' indicates that learning did not converge.
  Cases~\ref{MycounterA:alpha=1,beta=1} and \ref{MycounterA:alpha=0,beta=0} are equivalent to not including the  regularization method (we call this Vanilla) and RandomDrop, respectively.
  }
  \vspace{4pt}
    \label{tab:param_range}
    \begin{tabular}{c|c|c||c|c|c||c} \hline
    Case & $\alpha$ & $\beta$ & ResNet& ResNet (EraseReLU) & PyramidNet & Note \\ \hline \hline 
      \useMycounterA{\label{MycounterA:alpha=1,beta=1}} &1 & 1 & 27.42&25.38& 18.01 & Vanilla \\ \hline
      \useMycounterA{\label{MycounterA:alpha=0,beta=0}} &0 & 0 &24.07&22.86& 17.74 & RandomDrop\\ \hline \hline
      \useMycounterA{\label{MycounterA:alpha=1,beta=0}} &1 & 0 &27.95&27.57& 20.87 & \\ \cline{1-6}
      \useMycounterA{\label{MycounterA:alpha=1,beta=[0,1]}} &1 & $[0,1]$ &26.66&25.98& 18.80 & \\ \cline{1-6}
      \useMycounterA{\label{MycounterA:alpha=1,beta=[-1,1]}} &1 & $[-1,1]$ &28.45&28.23& 21.69 & \\ \cline{1-6}
      \useMycounterA{\label{MycounterA:alpha=0,beta=1}} &0 & 1 &27.15&39.09& \xmark & \\ \cline{1-6}
      \useMycounterA{\label{MycounterA:alpha=0,beta=[0,1]}} &0 & $[0,1]$ &\textbf{23.77}&\textbf{21.81}& \xmark & \\ \cline{1-6}
      \useMycounterA{\label{MycounterA:alpha=0,beta=[-1,1]}} &0 & $[-1,1]$ &24.69&23.22& \xmark & \\ \cline{1-6}
      \useMycounterA{\label{MycounterA:alpha=[0,1],beta=1}} & $[0,1]$ & $1$ &25.11&23.24& 38.48 \\ \cline{1-6}
      \useMycounterA{\label{MycounterA:alpha=[0,1],beta=0}} & $[0,1]$ & $0$ &25.93&24.73& 19.68 \\ \cline{1-6}
      \useMycounterA{\label{MycounterA:alpha=[0,1],beta=[0,1]}} & $[0,1]$ & $[0,1]$ &24.78&23.75& 18.27 \\ \cline{1-6}
      \useMycounterA{\label{MycounterA:alpha=[0,1],beta=[-1,1]}} & $[0,1]$ & $[-1,1]$ &26.41&25.32& 20.61 \\ \cline{1-6}
      \useMycounterA{\label{MycounterA:alpha=[-1,1],beta=1}} & $[-1,1]$ & $1$ &\xmark&39.98& 18.68 \\ \cline{1-6}
      \useMycounterA{\label{MycounterA:alpha=[-1,1],beta=0}} & $[-1,1]$ & $0$ &24.83&23.25& \textbf{17.28} \\ \cline{1-6}
      \useMycounterA{\label{MycounterA:alpha=[-1,1],beta=[0,1]}} & $[-1,1]$ & $[0,1]$ &\xmark&\textbf{22.59}& \textbf{16.22} \\ \cline{1-6}
      \useMycounterA{\label{MycounterA:alpha=[-1,1],beta=[-1,1]}} & $[-1,1]$ & $[-1,1]$ &25.85&23.91& 18.26 \\  \hline
    \end{tabular}
\end{table*}

\begin{table}[tb]
  \centering
  \caption{\textbf{[Update rule of $\alpha$ and $\beta$]} Average top-1 errors (\%) of ``PyramidNet + ShakeDrop'' for four runs at the final (300th) epoch on the CIFAR-100 dataset.
  }
  \vspace{4pt}
    \label{tab:Ex2_1}
    \begin{tabular}{c|c|c||c|c} \hline
    $\alpha$ & $\beta$ & Level & ResNet  &PyramidNet \\ \hline \hline
    \multirow{4}{*}{$0$} & \multirow{4}{*}{$[0,1]$} &Batch &23.77&\multirow{4}{*}{-} \\ \cline{3-4}
    & & Image &99.00&  \\ \cline{3-4}
    & & Channel &66.30&  \\ \cline{3-4}
    & & Pixel &\textbf{23.74}&  \\ \hline \hline
    \multirow{4}{*}{$[-1,1]$} & \multirow{4}{*}{$[0,1]$} &Batch &\multirow{4}{*}{-}& 16.22 \\ \cline{3-3} \cline{5-5}
    & & Image && 16.04 \\ \cline{3-3} \cline{5-5}
    & & Channel && 16.12 \\ \cline{3-3} \cline{5-5}
    & & Pixel && \textbf{15.78} \\ \hline \hline
    \end{tabular}
\end{table}
\begin{table*}[tbp]
  \centering
  \caption{\textbf{[Combinations of $(\alpha, \beta)$]} Top-1 errors (\%) of ``PyramidNet + ShakeDrop'' at the final (300th) epoch on the CIFAR-100 dataset in the batch-level update rule.
  Combinations of $\alpha$ and $\beta$ used in each case are marked.
  Results with $*$ are quoted from Table~\ref{tab:param_range}.}
   \label{tab:param_combination}
  \begin{tabular}{c||c|c|c|c||c||c} \hline
      Case & $\alpha=1,\beta=1$ & $\alpha=1,\beta=0$ & $\alpha=-1,\beta=1$ & $\alpha=-1,\beta=0$ & PyramidNet & Note \\ \hline \hline
      \useMycountera{\label{Mycountera:1111}} & \checkmark & \checkmark & \checkmark & \checkmark & \textbf{16.62} & \\ \hline \hline
     \useMycountera{\label{Mycountera:0111}} & & \checkmark & \checkmark & \checkmark & 19.51 &\\ \hline
     \useMycountera{\label{Mycountera:1011}} & \checkmark & & \checkmark & \checkmark & 18.79 &\\ \hline
     \useMycountera{\label{Mycountera:1101}} & \checkmark & \checkmark & & \checkmark & \textbf{16.57} &\\ \hline
     \useMycountera{\label{Mycountera:1110}} & \checkmark & \checkmark & \checkmark & & \textbf{16.43} &\\ \hline \hline
     \useMycountera{\label{Mycountera:0011}} & & & \checkmark & \checkmark & 37.11 &\\ \hline
     \useMycountera{\label{Mycountera:0101}} & & \checkmark & & \checkmark & 17.49 &\\ \hline
     \useMycountera{\label{Mycountera:0110}} & & \checkmark & \checkmark & & 17.25 &\\ \hline
     \useMycountera{\label{Mycountera:1001}} & \checkmark & & & \checkmark & \textbf{16.24}& \\ \hline
     \useMycountera{\label{Mycountera:1010}} & \checkmark & & \checkmark & & 17.39 &\\ \hline
     \useMycountera{\label{Mycountera:1100}} & \checkmark & \checkmark & & & 19.18 &\\ \hline \hline
     \useMycountera{\label{Mycountera:0001}} & & & & \checkmark & 25.81 &\\ \hline
     \useMycountera{\label{Mycountera:0010}} & & & \checkmark & &  88.75 & \\ \hline 
     \useMycountera{\label{Mycountera:0100}} & & \checkmark & & & $*$20.87 & Case~\ref{MycounterA:alpha=1,beta=0} in Table~\ref{tab:param_range} \\ \hline
     \useMycountera{\label{Mycountera:1000}} & \checkmark & & & & $*$18.01 & Vanilla \\ \hline \hline
    \end{tabular}
\end{table*}

\begin{table*}[tbp]
  \centering
  \caption{\textbf{[Comparison on CIFAR datasets]} Top-1 errors (\%) on CIFAR datasets.
  This table shows the results of the original networks (left) and modified networks (right).
  Modified networks refer to  the ``EraseReLU''ed versions in (a) and (c) and networks in which BN was inserted at the end of residual branches in (b).
  In ShakeDrop, $\alpha=0,\beta\in[0,1]$ was used in the original networks and $\alpha\in[-1,1],\beta\in[0,1]$ was used in the modified networks.
  In both cases, $p_L=0.5$ and the pixel-level update rule were used.
  ``\xmark'' indicates that learning did not converge.
    $*$ indicates that the result is quoted from the literature.
    $+$ indicates the average result of four runs.
  }
   \label{tab:error_architectures_CIFAR100}

\vspace{4pt}

   \subcaptionbox{Two-branch architectures (ResNet, ResNeXt, and PyramidNet) \label{tab:error_architectures_1branch_CIFAR100}}{
   \begin{tabular}{p{80mm}|c||c|c||c|c} \hline
      \multirow{2}{*}{Methods} & \multirow{2}{*}{Regularization} & \multicolumn{2}{c||}{CIFAR-10} & \multicolumn{2}{c}{CIFAR-100} \\ \cline{3-6}
       &  & Original & EraseReLU & Original & EraseReLU \\ \hline \hline
\multirow{3}{80mm}{\textbf{ResNet-110}\\ $<$Conv-BN-ReLU-Conv-BN-add-(ReLU)$>$} 
      & Vanilla &6.59&\xmark& $^+$27.42 & $^+$25.38 \\ \cline{2-6}
      & RandomDrop &5.51&5.16& $^+$24.07 & $^+$22.86 \\ \cline{2-6}
      & ShakeDrop &\textbf{4.56}&\textbf{4.81}& $^+$\textbf{23.74} & $^+$\textbf{21.81}\\ \hline \hline
\multirow{3}{80mm}{\textbf{ResNet-164 Bottleneck}\\ $<$Conv-BN-ReLU-Conv-BN-ReLU-Conv-BN-add-(ReLU)$>$} 
      & Vanilla &5.54&\xmark& 22.00 & 21.96 \\ \cline{2-6}
      & RandomDrop &5.27&4.71& 21.96 & 20.35\\ \cline{2-6}
      & ShakeDrop &\textbf{4.34}&\textbf{4.26}& \textbf{21.62} & \textbf{19.58}\\ \hline \hline
\multirow{3}{80mm}{\textbf{ResNeXt-29 8-64d Bottleneck}\\ $<$Conv-BN-ReLU-Conv-BN-ReLU-Conv-BN-add-(ReLU)$>$} 
      & Vanilla &4.79&4.75& 20.90 & 20.25 \\ \cline{2-6}
      & RandomDrop &\textbf{4.38}&4.9& \textbf{20.66} & 20.28\\ \cline{2-6}
      & ShakeDrop &\textbf{4.38}&\textbf{3.86}& 20.71 & \textbf{18.66}\\ \hline \hline
\multirow{3}{80mm}{\textbf{PyramidNet-110 $\alpha$270}\\ $<$BN-Conv-BN-ReLU-Conv-BN-add$>$} 
      & Vanilla &\multicolumn{2}{c||}{3.85}& \multicolumn{2}{c}{$^+$18.01} \\ \cline{2-6}
      & RandomDrop &\multicolumn{2}{c||}{3.63}& \multicolumn{2}{c}{$^+$17.74} \\ \cline{2-6}
      & ShakeDrop &\multicolumn{2}{c||}{\textbf{3.33}}& \multicolumn{2}{c}{$^+$\textbf{15.78}} \\ \hline
\multirow{3}{80mm}{\textbf{PyramidNet-272 $\alpha$200 Bottleneck}\\ $<$BN-Conv-BN-ReLU-Conv-BN-ReLU-Conv-BN-add$>$} 
      & Vanilla &\multicolumn{2}{c||}{3.53}& \multicolumn{2}{c}{$^*$16.35} \\ \cline{2-6}
      & RandomDrop &\multicolumn{2}{c||}{3.41}& \multicolumn{2}{c}{15.94} \\ \cline{2-6}
      & ShakeDrop &\multicolumn{2}{c||}{\textbf{3.08}} & \multicolumn{2}{c}{\textbf{14.96}} \\ \hline
    \end{tabular}
    }

\vspace{3mm}
   \subcaptionbox{Two-branch architectures (Wide-ResNet) \label{tab:error_architectures_1branch_wideresnet_CIFAR100}}{
   \begin{tabular}{p{85mm}|c||c|c||c|c} \hline
      \multirow{2}{*}{Methods} & \multirow{2}{*}{Regularization} & 
      \multicolumn{2}{c||}{CIFAR-10} & \multicolumn{2}{c}{CIFAR-100} \\ \cline{3-6}
       &  & Original & with BN & Original & with BN \\ \hline \hline
\multirow{3}{85mm}{\textbf{Wide-ResNet-28-10k}\\ $<$BN-ReLU-Conv-BN-ReLU-Conv-(BN)-add$>$} 
      & Vanilla &\textbf{4.05}&3.98& 20.67 & 20.05 \\ \cline{2-6}
      & RandomDrop &4.15&\textbf{3.85}& \textbf{19.95} & \textbf{19.16} \\ \cline{2-6}
      & ShakeDrop &87.6&4.37& 98.29  & 19.47\\ \hline
      \end{tabular}
}

\vspace{3mm}

   \subcaptionbox{Three-branch architectures\label{tab:error_architectures_2branch_CIFAR100}}{
    \begin{tabular}{p{75mm}|c||c|c||c|c} \hline
    \multirow{2}{*}{Methods} & \multirow{2}{*}{Regularization} & 
      \multicolumn{2}{c||}{CIFAR-10} & \multicolumn{2}{c}{CIFAR-100} \\ \cline{3-6}
       &  & Original & EraseReLU & Original & EraseReLU \\ \hline \hline
      \multirow{6}{75mm}{\textbf{ResNeXt-164 2-1-40d Bottleneck}\\ $<$Conv-BN-ReLU-Conv-BN-ReLU-Conv-BN-add-(ReLU)$>$} 
      & Vanilla &6.92&\xmark& 23.82 & 21.75 \\ \cline{2-6}
      & RandomDrop Type-A &5.00&\xmark& 21.38 & 20.44\\ \cline{2-6}
      & RandomDrop Type-B &6.78&\textbf{4.58}& \textbf{21.34} & 20.21\\ \cline{2-6}
      & Shake-Shake &5.61&4.65& 22.35 & 22.51 \\ \cline{2-6}
      & ShakeDrop Type-A &4.67&\xmark& 21.41 & 19.19\\ \cline{2-6}
      & ShakeDrop Type-B &\textbf{4.33}&\xmark& 21.52 & \textbf{18.66}\\ \hline \hline
      \multirow{6}{75mm}{\textbf{ResNeXt-29 2-4-64d Bottleneck}\\ $<$Conv-BN-ReLU-Conv-BN-ReLU-Conv-BN-add-(ReLU)$>$} 
      & Vanilla &5.39&5.01&21.19& \xmark \\ \cline{2-6}
      & RandomDrop Type-A &4.6&5.12& 21.12 & 20.13\\ \cline{2-6}
      & RandomDrop Type-B &4.13&4.42& 19.27 & 19.01\\ \cline{2-6}
      & Shake-Shake &4.64&3.84& 19.16 & 18.82 \\ \cline{2-6}
      & ShakeDrop Type-A &9.38&4.35& 22.51 & 18.49\\ \cline{2-6}
      & ShakeDrop Type-B &\textbf{3.91}&\textbf{3.67}& \textbf{18.27} & \textbf{17.80}\\ \hline
    \end{tabular}
}
\end{table*}

\begin{table*}[tbp]
  \centering
  \caption {\textbf{[Comparison on ImageNet]} Top-1 errors (\%) on ImageNet.
    This table shows the results of the original networks (left) and modified networks in which BN is at the end of the residual block (right).
  In ShakeDrop, $\alpha=0,\beta\in[0,1]$ were used in the original networks and $\alpha\in[-1,1],\beta\in[0,1]$ in the modified networks.
  In both cases, $p_L=0.9$ and the pixel-level update rule were used.
  }

   \label{tab:imagenet}
   \begin{tabular}{p{90mm}|c||c|c} \hline
      Methods & Regularization &Original & EraseReLU \\ \hline \hline
\multirow{3}{90mm}{\textbf{ResNet-152}\\ $<$Conv-BN-ReLU-Conv-BN-ReLU-Conv-BN-add-(ReLU)$>$} & Vanilla &21.72&22.79 \\ \cline{2-4}
      & RandomDrop &21.33&22.14   \\ \cline{2-4}
      & ShakeDrop &\bf{20.88}&\bf{21.78} \\ \hline \hline
\multirow{3}{90mm}{\textbf{ResNeXt-152}\\ $<$Conv-BN-ReLU-Conv-BN-ReLU-Conv-BN-add-(ReLU)$>$} & Vanilla &20.49& 22.57  \\ \cline{2-4}
      & RandomDrop&20.45 &22.09 \\ \cline{2-4}
      & ShakeDrop &\bf{20.34}&\bf{21.52}\\ \hline \hline
\multirow{3}{90mm}{\textbf{PyramidNet-152 $\alpha$300}\\ $<$BN-Conv-BN-ReLU-Conv-BN-ReLU-Conv-BN-add$>$} & Vanilla &\multicolumn{2}{c}{21.54 } \\ \cline{2-4}
      & RandomDrop&\multicolumn{2}{c}{21.23 }  \\ \cline{2-4}
      & ShakeDrop &\multicolumn{2}{c}{\bf{ 20.94} }  \\ \hline \hline
    \end{tabular}
\end{table*}

\begin{table}[tb]
  \centering
  \caption{
  \textbf{[Comparison on COCO datasets]} Average precision (\%) on the COCO minival dataset. 
  ``Det.'' denotes the average precision of object detection and ``Seg.'' denotes the average precision of instance segmentation.  
  }
  \label{tbl:coco}
  \vspace{4pt}
    \label{tab:COCO}
    \begin{tabular}{l|c||c|c} \hline
    Method                           &Regularization& Det.    &Seg. \\ \hline \hline
    \multirow{2}{*}{Faster R-CNN}     &Vanilla       &39.0     &-  \\ \cline{2-4}
    \multirow{2}{*}{(with ResNet-152)}&RandomDrop    &39.7     &-  \\ \cline{2-4}
                                     &ShakeDrop     &\bf{40.1}&-  \\ \hline \hline
    \multirow{2}{*}{Mask R-CNN}       &Vanilla       &39.7     &35.8 \\ \cline{2-4}	
    \multirow{2}{*}{(with ResNet-152)}&RandomDrop    &40.8     &36.6 \\ \cline{2-4}
                                     &ShakeDrop     &\bf{40.9}&\bf{36.9} \\ \hline \hline
    \end{tabular}
\end{table}

\begin{table}[tbp]
    \centering
    \caption{\textbf{[mixup + ShakeDrop]} Error rates (\%) of mixup + ShakeDrop.}
    \label{tbl:mixup+ShakeDrop}
    \begin{tabular}{c|l||c|c} \hline
        & Method & mixup & mixup + ShakeDrop \\ \hline \hline
        \multirow{7}{*}{CIFAR-100} & ResNet-110 & \textbf{23.79} & 24.12\\ \cline{2-4}
        & ResNet-164 & 22.93 & \textbf{21.13} \\ \cline{2-4}
        & ResNeXt-29 8-64d & 20.46 & \textbf{17.52} \\ \cline{2-4}
        & PyramidNet-110 $\alpha$270 & 17.47&\textbf{15.04} \\ \cline{2-4}
        & PyramidNet-272 $\alpha$200 & 16.61&\textbf{14.90} \\ \cline{2-4}
        & ResNeXt-164 2-1-40d & 22.13 & \textbf{20.97} (Type B) \\ \cline{2-4}
        & ResNeXt-29 2-4-64d & 20.54 & \textbf{17.10}  (Type B) \\ \hline \hline
         \multirow{1}{*}{ImageNet} & ResNet-152 & 21.46 & \textbf{21.15}\\ \hline
    \end{tabular}
\end{table}

\begin{table*}[tbp]
  \centering
  \caption{\textbf{[$p_L$ and depth]} Top-1 errors (\%) of ``ResNet + ShakeDrop'' and ``PyramidNet + ShakeDrop'' at the final (300th) epoch. 
  In ShakeDrop, $\alpha=0,\beta\in[0,1]$ were used in ResNet and $\alpha\in[-1,1],\beta\in[0,1]$ in PyramidNet.
  CIFAR-100 dataset in the channel-level update rule.
  $+$ indicates the average result of four runs.}
    \label{tab:pl_depth}
    \begin{tabular}{c|c|c||c|c|c|c|c|c} \hline
    Methods & Regularization & $p_L$  & 20 Layer & 38 Layer & 56 Layer & 74 Layer & 92 Layer & 110 Layer\\ \hline \hline
    \multirow{11}{*}{ResNet} & Vanilla
               & -     &  31.76   &  29.03   &  28.39   &  27.36   &  26.28   &  $^+$27.42 \\  \cline{2-9}
    &\multirow{5}{*}{RandomDrop}
             & 0.9   &  30.48   &  27.49   &  26.37   &  25.90   &  25.36   &  25.28     \\  \cline{3-9}
            && 0.8   &\textbf{30.39}&\textbf{26.65}&  26.14   &  24.92   &  25.50   &  24.54     \\  \cline{3-9}
            && 0.7   &  30.80   &  27.29   &\textbf{25.18}&\textbf{24.50}&\textbf{24.70}&\textbf{23.80} \\  \cline{3-9}
            && 0.6   &  32.61   &  26.97   &  25.88   &  24.67   &  24.77   &  24.39     \\  \cline{3-9}
            && 0.5   &  33.53   &  28.27   &  26.48   &  24.97   &  24.98   &  $^+$24.07 \\  \cline{2-9}
    &\multirow{5}{*}{ShakeDrop}
             & 0.9   &\textbf{29.66}&\textbf{26.81}&  26.26   &  25.73   &  25.50   &  24.38     \\  \cline{3-9}
            && 0.8   &  30.97   &  26.93   &  26.00   &  25.26   &  25.09   &  24.28     \\  \cline{3-9}
            && 0.7   &  32.93   &  27.01   &  26.14   &  24.83   &  24.88   &  23.99     \\  \cline{3-9}
            && 0.6   &  33.47   &  27.40   &\textbf{25.26}&  24.73   &  24.54   &  24.21     \\  \cline{3-9}
            && 0.5   &  36.09   &  28.56   &  26.00   &\textbf{24.57} &\textbf{23.98}&\textbf{$^+$23.74}\\ \hline \hline
            
    \multirow{11}{*}{PyramidNet} & Vanilla
               & -     &  22.52   &  19.89   &  19.37   &  18.41   &  18.67   &  $^+$18.01 \\  \cline{2-9}
    &\multirow{5}{*}{RandomDrop}
             & 0.9   &  21.00   &\textbf{18.75}&  18.71   &  17.59   &  17.70   &\textbf{16.97} \\  \cline{3-9}
            && 0.8   &\textbf{20.88}&  19.27   &  18.06   &\textbf{17.49}&  17.92   &  17.19     \\  \cline{3-9}
            && 0.7   &  21.08   &  19.02   &  18.47   &  17.89   &\textbf{17.62}&  17.20     \\  \cline{3-9}
            && 0.6   &\textbf{20.88}&  19.69   &\textbf{17.89}&  17.94   &  17.74   &  17.50     \\  \cline{3-9}
            && 0.5   &  21.88   &  19.87   &  18.52   &  18.06   &  17.69   &  $^+$17.74 \\  \cline{2-9}
    &\multirow{5}{*}{ShakeDrop}
             & 0.9   &  20.40   &  18.39   &  17.70   &  17.42   &  17.62   &  17.15     \\  \cline{3-9}
            && 0.8   &  20.36   &  18.57   &  17.97   &  16.96   &  17.38   &  16.69     \\  \cline{3-9}
            && 0.7   &  20.96   &\textbf{17.90}&  17.08   &  16.69   &  16.56   &  16.34     \\  \cline{3-9}
            && 0.6   &\textbf{20.29}&  18.17   &\textbf{17.07}&\textbf{16.50}&\textbf{16.42}&  16.46     \\  \cline{3-9}
            && 0.5   &  20.54   &  17.93   &  17.35   &  16.73   &  16.47   &\textbf{$^+$15.78}\\  \hline \hline
    \end{tabular}
\end{table*}

\section{Preliminary Experiments}
\label{sec:preliminary_experiments}

ShapeDrop has three parameters: $\alpha$, $\beta$, and $p_L$.
Additionally, four possible update rules of $\alpha$ and $\beta$ exist.
In this section, we search  for the best parameters of $\alpha$ and $\beta$ and best update rule on the CIFAR-100 dataset.
The best parameters found are used in the experiments in Section~\ref{sec:experiments}.
Following RandomDrop regularization~\cite{DBLP:conf/eccv/HuangSLSW16}, we used $p_L=0.5$ as the default.

\subsection{Ranges of $\alpha$ and $\beta$}
\label{subsec:range}

The best parameter ranges of $\alpha$ and $\beta$ were experimentally explored.
We applied ShakeDrop to three network architectures: ResNet, ResNet (EraseReLU version), and PyramidNet.
In the \textit{EraseReLU version}, the rectified linear unit (ReLU) at the bottom of the building blocks was erased~\cite{Dong_arXiv2017}.
Note that EraseReLU does not affect PyramidNet because it does not have the ReLU at the bottom of the building blocks.

Table~\ref{tab:param_range} shows the representative parameter ranges of $\alpha$ and $\beta$ that we tested and their results.
Cases \ref{MycounterA:alpha=1,beta=1} and \ref{MycounterA:alpha=0,beta=0} correspond to the vanilla network (i.e., without regularization) and RandomDrop, respectively.
On all three network architectures, case \ref{MycounterA:alpha=0,beta=0} was better than case \ref{MycounterA:alpha=1,beta=1}.
We consider the results of the three network architectures individually.
\begin{itemize}
    \vspace{-0.2\baselineskip}
    \setlength{\itemsep}{1pt}
    \item \textbf{PyramidNet} achieved the lowest error rates among the three network architectures.
    Only cases \ref{MycounterA:alpha=[-1,1],beta=0} and \ref{MycounterA:alpha=[-1,1],beta=[0,1]} outperformed case \ref{MycounterA:alpha=0,beta=0}.
    Among them, case \ref{MycounterA:alpha=[-1,1],beta=[0,1]} was the best.

    \item \textbf{ResNet} had a different tendency from PyramidNet:
    case \ref{MycounterA:alpha=[-1,1],beta=[0,1]}, which was the best on PyramidNet, did not converge.
    Only case \ref{MycounterA:alpha=0,beta=[0,1]} outperformed case \ref{MycounterA:alpha=0,beta=0}.

    \item \textbf{ResNet (EraseReLU version)} had the characteristics of both PyramidNet and ResNet;
    that is, both cases \ref{MycounterA:alpha=[-1,1],beta=[0,1]} and \ref{MycounterA:alpha=0,beta=[0,1]} outperformed case \ref{MycounterA:alpha=0,beta=0}.
    Case \ref{MycounterA:alpha=0,beta=[0,1]} was the best.
    \vspace{-0.2\baselineskip}
\end{itemize}
Through experiments using various base network architectures shown in Section~\ref{sec:experiments}, we found that case \ref{MycounterA:alpha=[-1,1],beta=[0,1]} was effective on ``EraseReLU''ed architectures.
By contrast, case \ref{MycounterA:alpha=0,beta=[0,1]} was effective on non-``EraseReLU''ed architectures.

\subsection{Update rule of $\alpha$ and $\beta$}
The best update rule of $\alpha$ and $\beta$ was found from the \textit{batch}, \textit{image}, \textit{channel}, and \textit{pixel} levels.
ShakeDrop is determined to drop or not on each building block. Differently from Dropout and RandomDrop, even if a building block is determined to be dropped, we still have a freedom to choose how $\alpha$ and $\beta$ are determined. That is, $\alpha$ and $\beta$ can be drawn for each batch in parallel, each image on a batch in parallel, each channel, or each element.

In the experiment, the best $\alpha$ and $\beta$ found in Section~\ref{subsec:range} (i.e., $\alpha=0$ and $\beta \in [0,1]$ for ResNet, and $\alpha \in [-1,1]$ and $\beta \in [0,1]$ for PyramidNet), were used%
\footnote{ Table~\ref{tab:Ex2_1} contains cells with a hyphen (``-''). We did not conduct experiments in these settings because they were not expected to improve the error rates in the experiment, as shown in Section~\ref{subsec:range}.}.
Table~\ref{tab:Ex2_1} shows that the pixel level was the best for both ResNet and PyramidNet.

\subsection{Combinations of $(\alpha, \beta)$ for analyzing ShakeDrop behavior}
Although we successfully found effective ranges of $\alpha$ and $\beta$, and their update rule, we still do not understand what mechanism contributes to improving the generalization performance of ShakeDrop.
One reason is that $\alpha$ and $\beta$ are random variables.
Because of  this, at the end of training, we can obtain a network that is trained using various observed values of $\alpha$ and $\beta$.
This makes it more difficult to understand the mechanism. 

Hence, in this section, we explore effective combinations of $(\alpha, \beta)$.
The combinations of $(\alpha, \beta)$ are defined as follows:
From the best ranges of $\alpha$ and $\beta$ for PyramidNet, which are $\alpha \in [-1, 1]$ and $\beta \in [0, 1]$, by taking both ends of ranges, we obtain a set of $(\alpha, \beta)$ pairs: $\{(1,1),(1,0),(-1,1),(-1, 0)\}$.
Then, we examine all its combinations, which are shown in Table~\ref{tab:param_combination}.
Intuitively, when $b_l=0$, instead of drawing $\alpha$ and $\beta$ in the ranges, a pair $(\alpha, \beta)$ is selected  from its pool with equal probability.

Table~\ref{tab:param_combination} shows combinations of $(\alpha, \beta)$ and their results on PyramidNet.
Compared with the best result in Table~\ref{tab:param_range} (i.e., case \ref{MycounterA:alpha=[-1,1],beta=[0,1]}; 16.22\%), the results in Table~\ref{tab:param_combination} are almost comparable.
In particular, the best result in Table~\ref{tab:param_combination} (i.e., case \ref{Mycountera:1001}; 16.24\%) is almost equivalent.
This indicates that the random drawing of $\alpha$ and $\beta$ in certain ranges is not the primary factor for improving error rates.

Additionally, we observe that $p_L$ is important to error rates.
    As mentioned above, $(1,1)$ is the normal state.
    Hence, the difference between cases \ref{Mycountera:1001} and \ref{Mycountera:0001} exists only in $p_L$:
    because case \ref{Mycountera:1001} has two elements and one of them is the normal state (i.e., $(1,1)$), its $p_L$ actually works as  $(1+p_L) / 2$.
    For example, when $p_L=0.5$, case \ref{Mycountera:1001} is equivalent to case \ref{Mycountera:0001} with $p_L=0.75$.
    Cases \ref{Mycountera:1010} and \ref{Mycountera:0010}, and cases \ref{Mycountera:1100} and \ref{Mycountera:0100} have the same relationship.
    A comparison of their error rates shows that $p_L$ greatly affects the error rates.
    We discuss this issue in Section~\ref{sec:P_L_sensitivity}.

For further analysis, we focus on the difference among $(1,0)$, $(-1,1)$, and $(-1,0)$.
\vspace{0.5\baselineskip}

\noindent
\textbf{What does $(\alpha, \beta)=(1,0)$ do? (What the meaning  of $\beta=0$?)}\\
    $\alpha=1$ indicates that the forward pass is normal.
    Hence, no regularization effect is expected.
    When $\beta=0$, the network parameters of \textit{the layers selected for perturbat ion} (i.e., the layers with $b_l=0$) are not updated.
    In layers other than the selected layers, the network parameters are updated as usual.
    One exception is that, as the network parameters of the selected layers are not updated, other layers compensate for the amount that should be updated on the selected layers.
    Cases \ref{Mycountera:1100} and \ref{Mycountera:0100} contain $(1,0)$.
    They were slightly worse than the best cases .
\vspace{0.5\baselineskip}

\noindent
\textbf{What does $(\alpha, \beta)=(-1,1)$ do? (What the meaning of $\alpha=-1$?)}\\
    When $\alpha=-1$, in the selected layers, the calculation of the forward pass is perturbed by $\alpha=-1$.
    Then, the effect of perturbation is propagated to the succeeding layers.
    Hence, not only the selected layers but also their succeeding layers are perturbed.
    In the backward pass, when $\alpha$ is negative, the network parameters of the selected layers are updated toward the opposite direction to usual.
    Because of this, the network parameters of the selected layers are strongly perturbed by negative $\alpha$.
    This can be a destructive update.
    In layers other than the selected layers, it is less probable that the update of the network parameters is destructive
    because they follow the normal update rule (equivalent to $\alpha=1$).
    Cases \ref{Mycountera:1010} and \ref{Mycountera:0010} contain $(-1,1)$.
    The former was slightly worse than the best and the latter was significantly bad.
\vspace{0.5\baselineskip}

\noindent
\textbf{What does $(\alpha, \beta)=(-1,0)$ do?}\\
    As this is a combination of $\alpha=-1$ and $\beta=0$, their combined effect occurs.
    Following the case of $\alpha=-1$ mentioned above, the calculation of the forward pass is perturbed, and its effect is propagated to the succeeding layers.
    In the backward pass, following the case of $\beta=0$, the network parameters of only the selected layers are not updated.
    This can avoid \textit{destructive} updates caused by negative $\alpha$.
    Hence, $(-1,0)$ is expected to be effective.
    Cases \ref{Mycountera:1001} and \ref{Mycountera:0001} contained $(-1,0)$, and the former was the best.
\vspace{0.5\baselineskip}

\noindent
By extending the discussion above, we can interpret the behavior of ShakeDrop using $\alpha=0$ and $\beta \in [0,1]$, which was the most effective on ResNet.
When $\alpha=0$, in the forward pass, the outputs of the selected layers are identical to the inputs.
In the backward pass, the amount of updating of the network parameters is perturbed by $\beta$.

\section{Experiments}
\label{sec:experiments}

\subsection{Comparison on CIFAR datasets}
\label{sec:experiments_cifar100}

The proposed ShakeDrop was compared with RandomDrop and Shake-Shake in addition to the vanilla network (without regularization) on ResNet, Wide ResNet, ResNeXt, and PyramidNet.
Implementation details are available in Appendix~\ref{sec:implement}.

Table~\ref{tab:error_architectures_CIFAR100} shows the conditions and experimental results on CIFAR datasets~\cite{CIFAR-10_100}.
In the table, method names are followed by the components of their building blocks.
We used the parameters of ShakeDrop found in Section~\ref{sec:preliminary_experiments};
that is, the original networks used $\alpha=0,\beta\in[0,1]$ and the modified networks in which the residual branches end with BN (e.g., EraseReLU versions) used $\alpha\in[-1,1],\beta\in[0,1]$.
In ResNet and two-branch ResNeXt, in addition to the original form, EraseReLU versions
were examined.
In Wide ResNet, BN was added to the end of residual branches so that the residual branches ended with BN. 
In three-branch ResNeXt, we examined two approaches, referred to as ``Type A'' and ``Type B,'' to apply RandomDrop and ShakeDrop.
``Type A'' and ``Type B'' indicate that the regularization unit was inserted after and before the addition unit for residual branches, respectively;
that is, on the forward pass of the training phase, Type A is given by
\begin{equation}
\label{eqn:typea}
    G(x) = x + D(F_1(x)+F_2(x)),
\end{equation}
where $D(\cdot)$ is a perturbation unit of RandomDrop or ShakeDrop, and
Type B is given by
\begin{equation}
\label{eqn:typeb}
    G(x) = x + D_1(F_1(x))+D_2(F_2(x)), 
\end{equation}
where $D_1(\cdot)$ and $D_2(\cdot)$ are individual perturbation units.

Table~\ref{tab:error_architectures_CIFAR100} shows that ShakeDrop can be applied not only to three-branch architectures (ResNeXt) but also two-branch architectures (ResNet, Wide ResNet, and PyramidNet), and ShakeDrop outperformed RandomDrop and Shake-Shake, except for some cases.
In Wide ResNet with BN, although ShakeDrop improved the error rate compared with the vanilla network, it did not compared with RandomDrop.
This is because the network only had 28 layers.
As shown in the RandomDrop paper~\cite{DBLP:conf/eccv/HuangSLSW16}, RandomDrop is less effective on a shallow network and more effective on a deep  network.
We observed the same phenomenon in ShakeDrop, and ShakeDrop is more sensitive than RandomDrop.
See Section~\ref{sec:P_L_sensitivity} for more detail.

\subsection{Comparison on the ImageNet dataset}
\label{subsec:imagenet}

We also conducted experiments on the ImageNet classification dataset~\cite{Krizhevsky_NIPS2012} using ResNet, ResNeXt, and PyramidNet of 152 layers.
The implementation details are presented in Appendix~\ref{sec:implement}.
We used the best parameters found on the CIFAR datasets, except for $p_L$.
We experimentally selected $p_L=0.9$.

Table~\ref{tab:imagenet} shows the experimental results.
Contrary to the CIFAR cases, the EraseReLU versions were worse than the original networks, which does not support the claim of the EraseReLU paper~\cite{Dong_arXiv2017}.
On ResNet and ResNeXt, in both the original and EraseReLU versions, ShakeDrop clearly outperformed RandomDrop and the vanilla network (ShakeDrop gained 0.84\% and 0.15\% compared with the vanilla network in the original networks, respectively).
On PyramidNet, ShakeDrop outperformed the vanilla network
(ShakeDrop gained 0.60\% compared with the vanilla network) and also RandomDrop (ShakeDrop gained by 0.29\% compared with RandomDrop).
Therefore, on ResNet, ResNeXt, and PyramidNet, ShakeDrop clearly outperformed RandomDrop and the vanilla network.

\subsection{Comparison on the COCO dataset}
\label{subsec:coco}

From the results in Sections~\ref{sec:experiments_cifar100} and  ~\ref{subsec:imagenet}, we considered that ShakeDrop promoted the generality of feature extraction and we evaluated the generality on the COCO dataset~\cite{DBLP:conf/eccv/Tsung14}.
We used Faster R-CNN and Mask R-CNN with the ImageNet pre-trained original version ResNet of 152 layers in Section~\ref{subsec:imagenet}.
The implementation details are presented in Appendix~\ref{sec:implement}.

Table~\ref{tab:COCO} shows the experimental results. On Faster R-CNN and Mask R-CNN, ShakeDrop clearly outperformed RandomDrop and the vanilla network.
Therefore, ShakeDrop promoted the generality of feature extraction not only for image classification but also detection and instance segmentation.

\subsection{Simultaneous use of ShakeDrop with mixup}
\label{sec:mixup}

As mentioned in Section~\ref{subsec:relationship_existing_methods}, we have successfully used ShakeDrop combined with mixup.
Table.~\ref{tbl:mixup+ShakeDrop} shows the results.
In most cases, ShakeDrop further improved the error rates of the base neural networks to which mixup was applied.
This indicates that ShakeDrop is not a rival to other regularization methods, such as mixup, but a  ``collaborator.''

\subsection{Relationship between network depth and best $p_L$}
\label{sec:P_L_sensitivity}

As mentioned in Section~\ref{subsec:regularization}, it has been  experimentally found that RandomDrop is more effective on deeper networks (see the figure on the right in Fig. 8 of \cite{DBLP:conf/eccv/HuangSLSW16}).
We performed similar experiments on ShakeDrop and RandomDrop to compare their sensitivity to the depth of networks.

Table~\ref{tab:pl_depth} shows that the error rates varied over both $p_L$ and the network depth.
ShakeDrop with a large $p_L$ tended to be effective in shallower networks.
The same observation was obtained in the experimental study on the relationship between $p_L$ of RandomDrop and generalization performance~\cite{DBLP:conf/eccv/HuangSLSW16}.
We recommend a large  $p_L$ for shallower network architectures.

\section{Conclusion}
We proposed a new stochastic regularization method called ShakeDrop which, in principle, can be applied to the ResNet family.
Through experiments on the CIFAR and ImageNet datasets, we confirmed that, in most cases, ShakeDrop outperformed existing regularization methods of the same category, that is,  Shake-Shake and RandomDrop.

\appendices

\appendices
\begin{table*}[tbp]
    \centering
    \caption{\textbf{[Learning conditions on CIFAR datasets]} Learning conditions of the original experiments and our experiments.
    ``Init." denotes  the initial learning rates (CIFAR-10/CIFAR-100).
    ``Total" denotes the total epoch number and ``WD" denotes the weight decay.
    ``it." denotes iterations and ''ep." denotes epochs.
    ``-" indicates that a value was not specified  in the original paper.
    Bold items for our experiments indicate changes from the original conditions.
    We used the most common conditions for the original conditions, except for the initial learning rates and number of GPUs.
    Other than ResNet, the original learning rates were used in the experiments.
    On ResNet, the learning rate was 0.1 because our learning rate schedule did not warm up the training  for the first 0.4k iterations.
    We used four GPUs to accelerate learning as much as possible. 
    }
    \label{tab:main_change}
    \begin{tabular}{c|c||c|c|c|c|c} \hline
        Methods & Version & Init.&Learning rate schedule (operation timing)& Total & \#GPU & WD\\ \hline \hline
        \multirow{2}{*}{ResNet} & Original~\cite{DBLP:conf/eccv/HeZRS16}
        & 0.01 / 0.01&$\times 10$ (0.4k it.)$\to \times 0.1$ (32k it.)$\to \times 0.1$ (48k it.)
        & -   & 2& 0.0001\\ \cline{2-7}
        & Ours 
        & \textbf{0.1 / 0.1}& \textbf{{\boldmath$\times 0.1$} (150 ep.){\boldmath$\to \times 0.1$} (225 ep.)}
        & \textbf{300} & \textbf{4} & 0.0001 \\ \hline \hline
        \multirow{2}{*}{ResNeXt} & Original~\cite{Xie_CVPR2017}
        &  0.1 / 0.1&$\times 0.1$ (150 ep.)$\to \times 0.1$ (225 ep.)& 300 & 8 & 0.0005\\ \cline{2-7}
        & Ours 
        &  0.1 / 0.1&$\times 0.1$ (150 ep.)$\to \times 0.1$ (225 ep.)& 300 & \textbf{4} & \textbf{0.0001} \\ \hline \hline
        \multirow{2}{*}{PyramidNet} & Original~\cite{Han_CVPR2017}
        &  0.1 / 0.5&$\times 0.1$ (150 ep.)$\to \times 0.1$ (225 ep.)& 300 & - & 0.0001\\ \cline{2-7}
        & Ours 
        &  0.1 / 0.5&$\times 0.1$ (150 ep.)$\to \times 0.1$ (225 ep.)& 300 & \textbf{4} & 0.0001 \\ \hline \hline
        \multirow{2}{*}{Wide ResNet} & Original~\cite{Zagoruyko2016WRN}
        &  0.1 / 0.1&$\times 0.2$ (60 ep.)$\to \times 0.2$ (120 ep.)$\to \times 0.2$ (160 ep.)& 200 & 1 & 0.0005\\ \cline{2-7}
        & Ours 
        &  0.1 / 0.1&\textbf{{\boldmath$\times 0.1$} (150 ep.){\boldmath$\to \times 0.1$} (225 ep.)}& \textbf{300} & \textbf{4} & \textbf{0.0001} \\ \hline \hline

    \end{tabular}
\end{table*}

\section{Experimental Conditions}
\label{sec:implement}
All networks were trained using back-propagation by SGD with the Nesterov accelerated gradient~\cite{NAG} and momentum method~\cite{1986Natur.323..533R}.
Four GPUs (on CIFAR) and eight GPUs (on ImageNet) were used for learning acceleration: because of parallel processing, different observations of $b_l$, $\alpha$, and $\beta$ were obtained on each GPU.
For example, the $l$-th layer on a GPU could be perturbed, whereas the layer was not perturbed on other GPUs ($l$ is an arbitrary number). 
Additionally, even if the layer was perturbed on multiple GPUs, the different observations of $\alpha$ and $\beta$ could be used depending on each GPU.

All implementations used in the experiments were based on the publicly available code of ResNet\footnote{\url{https://github.com/facebook/fb.resnet.torch}}, ResNeXt\footnote{\url{https://github.com/facebookresearch/ResNeXt}}, PyramidNet\footnote{\url{https://github.com/jhkim89/PyramidNet}}, Wide ResNet\footnote{\url{https://github.com/szagoruyko/wide-residual-networks}},  Shake-Shake\footnote{\url{https://github.com/xgastaldi/shake-shake}}, and Faster/Mask R-CNN\footnote{\url{https://github.com/facebookresearch/maskrcnn-benchmark}}.
We changed their various learning conditions to make them as common as possible  on CIFAR (in Section~\ref{sec:experiments_cifar100}).
Table~\ref{tab:main_change} shows the main changes.
The implementation is available at \textit{\url{https://github.com/imenurok/ShakeDrop}}.

The experimental conditions for each type of dataset are described below. 
\vspace{0.5\baselineskip}

\noindent
\textbf{CIFAR datasets}
The input images of CIFAR datasets~\cite{CIFAR-10_100} were processed in the following manner.
The original images of $32 \times 32$ pixels were color-normalized and then horizontally flipped with a 50\% probability.
Then, they were zero-padded to be $40 \times 40$ pixels and randomly cropped to be images of $32 \times 32$ pixels.
%An input image is a resolution of $32 \times 32$ pixels randomly cropped from a zero-padded $40 \times 40$ color normalized image or its horizontal flipping.
On PyramidNet, the initial learning rate was set to 0.1 on CIFAR-10 and 0.5 on CIFAR-100 following the PyramidNet paper~\cite{Han_CVPR2017}.
Other than PyramidNet, the initial learning rate was set to 0.1.
The initial learning rate was decayed by a factor of 0.1 at 150 epochs and 225 epochs of the entire learning process (300 epochs), respectively.
Additionally, a weight decay of 0.0001, momentum of 0.9, and batch size of 128 were used on four GPUs.
``MSRA''~\cite{He_ICCV2015} was used as the filter parameter initializer.
We evaluated the top-1 errors without any ensemble technique.
Linear decay parameter $p_L = 0.5$ was used following the RandomDrop paper~\cite{DBLP:conf/eccv/HuangSLSW16}.
ShakeDrop used parameters of $\alpha=0, \beta=[0,1]$ (Original) and $\alpha=[-1,1], \beta=[0,1]$ (EraseReLU on ResNet and ResNeXt, Wide ResNet with BN, and PyramidNet) with the pixel-level update rule.
\vspace{0.5\baselineskip}

\noindent
\textbf{ImageNet dataset}
The input images of ImageNet~\cite{imagenet_cvpr09} were processed in the following manner.
The original image was distorted using a random aspect ratio~\cite{Szegedy_CVPR2015} and randomly cropped to an image size of $224 \times 224$ pixels.
Then, the image was horizontally flipped with a 50\% probability and standard color noise~\cite{Krizhevsky_NIPS2012} was added.
On PyramidNet, the initial learning rate was set to 0.5.
The initial learning rate was decayed by a factor of 0.1 at $60$, $90$, and $105$ epochs of the entire learning process (120 epochs) following \cite{Han_CVPR2017}.
Additionally, a batch size of 128 was used on eight GPUs.
Other than PyramidNet, the initial learning rate was set to 0.1.
The initial learning rate was decayed by a factor of 0.1 at $30$, $60$, and $80$ epochs of the entire learning process (90 epochs) following \cite{Goyal_arXiv2017}.
Additionally, a batch size of 256 was used on eight GPUs.
A weight decay of 0.0001 and momentum of 0.9 were used.
``MSRA''~\cite{He_ICCV2015} was used as the filter parameter initializer.
We evaluated the top-1 errors without any ensemble technique on the single $224 \times 224$ image that was cropped from the center of  an image resized with the shorter side $256$.
$p_L=0.9$ was used as the linear decay parameter.
ShakeDrop used parameters of $\alpha=0, \beta=[0,1]$ (Original) and $\alpha=[-1,1], \beta=[0,1]$ (EraseReLU on ResNet and ResNeXt, and PyramidNet) with the pixel-level update rule.
\vspace{0.5\baselineskip}

\noindent
\textbf{COCO dataset}
Input images of COCO~\cite{DBLP:conf/eccv/Tsung14} were processed in the following manner.
We trained models on the union of the 80k training set and 35k val subset, and evaluated the models on the remaining 5k val subset.
We used ResNet-152 for the backbone  network and FPN~\cite{Lin_CVPR2017} for the predictor network.
To use ResNet-152 as a feature extractor, we used the expected value $E(b_l + \alpha - b_l \alpha)$ instead of ShakeDrop regularization. 
According to the experimental condition of the ImageNet dataset, the original image was color-normalized with the means and standard deviations of ImageNet dataset images.
The initial learning rate was set to 0.2.
The initial learning rate was decayed by a factor of 0.1 at $60,000$ and $80,000$ iterations of the entire learning process (90,000 iterations).
Additionally, a batch size of 16 was used on eight GPUs.
A weight decay of 0.0001 was used.
The other experimental conditions were set according to \textit{maskrcnn-benchmark}\footnotemark[8].

\section*{Acknowledgment}
We thank Maxine Garcia, PhD, from Edanz Group (www.edanzediting.com/ac) for editing a draft of this manuscript.

\bibliography{egbib}
\bibliographystyle{ieee}

\EOD

\end{document}